\documentclass[runningheads]{llncs}
\usepackage[T1]{fontenc}
\usepackage{graphicx}
\usepackage{comment}
\usepackage{booktabs}
\usepackage{hyperref}
\usepackage{paralist}
\usepackage{pifont}% http://ctan.org/pkg/pifont
\newcommand{\checkmark}{\ding{51}}
\newcommand{\xmark}{\ding{53}}
\usepackage{xcolor}
\usepackage{makecell}
\usepackage{amsmath}
\usepackage{siunitx}
\usepackage{orcidlink}
\begin{document}
\title{Enhancing the Learning Experience: Using Vision-Language Models to Generate Questions for Educational Videos}
\titlerunning{Using VLMs to Generate Questions for Educational Videos}
\author{Markos Stamatakis\inst{1}\orcidlink{0000-0002-7974-308X} \and
Joshua Berger\inst{2}\orcidlink{0009-0001-0750-4143} \and
Christian Wartena\inst{2}\orcidlink{0000-0001-5483-1529} \and
Ralph Ewerth\inst{1,3,4}\orcidlink{0000-0003-0918-6297} \and
Anett Hoppe\inst{1,3,4}\orcidlink{0000-0002-1452-9509}
}
\authorrunning{M. Stamatakis et al.}
\institute{TIB -- Leibniz Information Centre for Science and Technology, Hannover, Germany
\and
Hochschule Hannover -- Data$|$H Institute for Applied Data Science, Hannover, Germany 
\and
L3S Research Center -- Leibniz University Hannover, Hannover, Germany
\and University of Marburg and hessian.AI -- Hessian Center for Artifical Intelligence, Marburg, Germany\\
%\and hessian.AI – Hessian Center for Artificial Intelligence, Germany\\
\email{markos.stamatakis@tib.eu}}
%\institute{Princeton University, Princeton NJ 08544, USA \and
%Springer Heidelberg, Tiergartenstr. 17, 69121 Heidelberg, Germany
\maketitle              % typeset the header of the contribution
\begin{abstract}
Web-based educational videos offer flexible learning opportunities and are becoming increasingly popular. 
However, improving user engagement and knowledge retention remains a challenge.
Automatically generated questions can activate learners and support their knowledge acquisition. 
Further, they can help teachers and learners assess their understanding. 
While large language and vision-language models have been employed in various tasks, their application to question generation for educational videos remains underexplored. 
In this paper, we investigate the capabilities of current vision-language models for generating learning-oriented 
questions for educational video content. %, with implications for knowledge representation and adaptive educational technologies.
We assess (1)~out-of-the-box models' performance; (2)~fine-tuning effects on content-specific question generation; (3)~the impact of different video modalities on question quality; and (4)~in a qualitative study, question relevance, answerability, and difficulty levels of generated questions. 
Our findings delineate the capabilities of current vision-language models, highlighting the need for fine-tuning and addressing challenges in question diversity and relevance.
We identify requirements for future multimodal datasets and outline promising research directions.
\end{abstract}



\section{Introduction}
\label{sec:intro}
The use of video content has grown across web platforms, transforming how information is shared and consumed in social media and for entertainment.
In education, videos offer unique advantages: they combine visual and auditory elements to enhance the learning experience~\cite{Leisner2020DifferentWO,self_paced_learning} while providing students with flexibility in when and where they study~\cite{MERKT2022104355}.
However, this flexibility and passive engagement of video watching can reduce active engagement.
Question generation addresses this challenge by prompting learners to process the content actively~\cite{engagment_learning}.
Automatically generating questions from the most relevant information in a video can enhance learner interaction and knowledge retention.
 
Before large language models (LLMs), Natural Language Processing (NLP) relied on rule-based systems, machine learning, and early neural networks~\cite{survey_nlp}.
However, recent research has shown that LLMs achieve groundbreaking results, significantly improving tasks like text summarization and translation.
Models like GPT-4~\cite{openai2023gpt4} (Generative Pre-trained Transformer) and vision-language models (VLMs), such as OpenFlamingo~\cite{awadalla2023openflamingo}, BLIP-2~\cite{Li2023BLIP2BL} and LLaVA~\cite{liu2023improvedllava,liu2023llava} can process images and text.
While video content analysis with these models is still in early stages, focusing mainly on monomodal, text-oriented approaches using speech transcripts~\cite{lin2023video,luo2023valley,Maaz2023VideoChatGPT}, some newer models can process videos multimodally~\cite{li2024videochat,munasinghe2023PGVideoLLaVA,damonlpsg2023videollama}.
We explore the potential of multimodal VLMs for educational video question generation 
using answer-unaware question generation.
We investigate how current models perform without training and whether fine-tuning with appropriate datasets can improve question quality.
Beyond standard automatic metrics, we introduce a custom metric that compares questions with speech transcripts to assess content alignment.
This paper examines these capabilities through the following research questions~(RQs):
\begin{itemize}
  \item[\textbf{RQ1:}] How well do VLMs generate learning-oriented questions from educational videos in zero-shot settings?
  \item[\textbf{RQ2:}] Does fine-tuning VLMs improve question quality?
  \item[\textbf{RQ3:}] Do existing datasets suffice for high-quality question generation?
  \item[\textbf{RQ4:}] What dataset characteristics are essential for training and evaluating question generation models?
\end{itemize}

We first present related work regarding question generation and useful datasets (Section~\ref{sec:rl}).
Then, we outline the used data and chosen models (Section~\ref{sec:comp_vlms}).
Later, we discuss our experiments and their results, including zero-shot, fine-tuning, an ablation study, and a statistical and manual qualitative evaluation (Section~\ref{sec:exp_evaluation}).
Finally, a conclusion is drawn (Section~\ref{sec:conclusion}).
\section{Related Work}
\label{sec:rl}
This section presents related work on
automatic question generation for educational tasks (Section~\ref{subsec:rl_qg}), and relevant datasets 
(Section~\ref{subsec:useful_data}). 
\subsection{Question Generation for Educational Material}
\label{subsec:rl_qg}
Recent research on educational question generation can be classified into two main approaches in our context: one focuses on monomodal data sources like text documents, and another focuses on 
multimodal data like videos.
Much research has been done on text-based approaches regarding whether LLMs are capable of a more learning-focused question generation. 
Specifically, the research explored to what extent LLMs can generate questions of various complexity levels and whether they can accurately classify a given question into the appropriate category~\cite{qg_graessers,qg_bloom2,qg_school_level,qg_bloom}.
Their focus is primarily on Bloom’s taxonomy~\cite{blooms_taxonomy}, while Graesser’s taxonomy~\cite{graessers_taxonomy} also appears in the categorization of questions.
Furthermore, studies have investigated if generated questions are meaningful in terms of learning goals~\cite{mcq_educator,relevance_edu_qg,topic_controlled_qg,qg_university,DBLP:conf/aied/ShimmeiBM23,medicalexam_qg}.
This is facilitated by defined metrics such as topic relevance, answerability, and learning utility.
It can be observed that, in addition to the standard generation of questions, multiple-choice questions are also a focus, requiring an additional evaluation of their answer options.

Looking at multimodal data, it becomes evident that research focuses more on using only one modality to create questions for this kind of data.
Especially for videos, the question generation mostly relies on speech transcripts~\cite{relevant_video_qg,ICWSM18LearningQ,videodl,skanda_qg}.
This comes with the drawback that the visual content of videos is ignored, preventing generating questions on this information.
\subsection{Relevant Datasets}
\label{subsec:useful_data}
While generic question-answering datasets, such as SQuAD~\cite{rajpurkar2016squad} and WikiQA~\cite{wikiQA}, are useful for training and evaluating question-generation models, they often lack the educational focus required for our study.
In contrast, datasets like EduQG~\cite{eduqg} are 
designed for educational purposes, providing multiple-choice questions and answers tailored to this domain.
These datasets focus especially on text-based data, making them suitable only for monomodal approaches.

When considering multimodal approaches, particularly those involving videos for our study, existing video datasets cover a wide range of domains, including TV series~\cite{lei2018tvqa,lei2019tvqa}, tutorial videos~\cite{tutorialvqa}, activities in web videos~\cite{yu2019activityqa}, medicine~\cite{gupta2023dataset}, and more general videos~\cite{xu2017video}.
In this context, \textit{LearningQ}~\cite{ICWSM18LearningQ} stands out as a dataset specifically designed for question generation in education, containing instructor-designed questions from TED-Ed and learner-generated questions from Khan Academy for corresponding videos and articles.

\textit{TQA}~\cite{tqa} is an educational dataset that collects online textbooks to facilitate multimodal question generation. It incorporates texts, diagrams, and images, with the current version also including videos and questions about the content.

To the best of our knowledge, no suitable dataset currently includes annotations for the temporal position of questions, 
necessitating the use of entire videos for question generation rather than relevant segments.
Additionally, a higher diversity of video presentation styles in a dataset would be beneficial for a more detailed analysis of model behavior.
\section{Datasets, Models, Prompt Design and Metrics}
\label{sec:comp_vlms}
Our review of educational question generation highlights a research gap concerning multimodal approaches and the investigation of the capabilities of state-of-the-art VLMs for generating questions from educational videos. 
We evaluate their performance with and without fine-tuning (RQ1, RQ2), assess the adequacy of current datasets (RQ3), and identify key dataset characteristics for effective training and evaluation (RQ4).
Furthermore, we explore different prompts and evaluation metrics to gain deeper insights into model behavior (RQ1, RQ2).

\subsection{Datasets}
\label{subsec:methodology_data}
We conducted our experiments using the \textit{LearningQ} dataset~\cite{ICWSM18LearningQ}, which provides quality-checked videos and annotated questions. 
We chose this dataset over TQA~\cite{tqa}, which has copyright restrictions preventing video access, and over text-only datasets like SQuAD~\cite{rajpurkar2016squad} unsuitable for our video-focused study.

\paragraph{Data Acquisition:}
We acquired the data from LearningQ and automatically downloaded Khan Academy and TED-Ed videos. 
Following the authors' recommendation~\cite{ICWSM18LearningQ}, we only used questions labeled as useful for learning.
Further, we excluded questions related to text documents.
For TED-Ed videos, we retained multiple-choice and open-ended questions but filtered cloze tests and indirect questions (without a question mark).
\paragraph{Data Statistics:}
\begin{table}[tb]
\centering
%\fontsize{8}{10}\selectfont
%\def\arraystretch{0.9}
\caption{Statistics of the used data before and after (indicated with *) pre-processing.}
\label{tab:used_data}
%\resizebox{\columnwidth}{!}{
\begin{tabular}{lcccc}
  \hline
  \textbf{Dataset} & \textbf{TED-Ed} & \textbf{TED-Ed*} & \textbf{Khan} & \textbf{Khan*}\\
  \hline
  \textbf{\#Videos} & \num{1482} & \num{1417} & \num{7925} & \num{7863} \\
  \textbf{\#Questions} & \num{9720} & \num{7007} & \num{204771} & \num{204213}  \\
  \textbf{\#Avg Questions} & \num{6.69} & \num{4.95} & \num{25.84} & \num{25.97} \\
  \textbf{Min Video Length} & - & 0:00:23 & - & 0:00:39 \\
  \textbf{Avg Video Length} & - & 0:06:02 & - & 0:07:08 \\
  \textbf{Max Video Length} & - & 0:27:04 & - & 1:12:27 \\
  \hline
\end{tabular}
%}
\end{table}
Table~\ref{tab:used_data} presents the dataset structure and size statistics after our pre-processing steps.
The final dataset differs from the original version of the authors' report~\cite{ICWSM18LearningQ} due to significant reductions: 
In the TED-Ed dataset, no questions were provided for \num{28} videos, and \num{37} additional videos were inaccessible. 
\num{7007} questions remained (\qty{72.09}{\percent}), reducing the average question per video from \num{6.69} to \num{4.95}.
For the Khan Academy dataset, \num{62} videos (\qty{0.82}{\percent}) could not be downloaded, resulting in \num{558} removed tasks (\qty{99.73}{\percent} retained).
Video lengths vary considerably.
Khan Academy videos can exceed an hour, while TED-Ed videos are limited to around \num{30} minutes.
However, most videos are much shorter.
\paragraph{Data Split:}
We employed an \num{80}/\num{10}/\num{10} split for training, validation, and testing, randomly distributing video IDs across these sets (seed value: \num{1234}). 
Table~\ref{tab:datasplit} summarizes the distribution of videos/questions across the splits.
Notably, the splits maintained an even balance between multiple-choice and open-ended questions, allowing for question-type-independent model training and evaluation. 
\begin{table}[tb]
\centering
%\fontsize{8}{10}\selectfont
%\def\arraystretch{0.9}
\caption{Pre-processed data statistics for train, val, and test sets. The first half represents the number of videos and the second half represents the number of questions.}
\label{tab:datasplit}
%\resizebox{\columnwidth}{!}{
\begin{tabular}{lccc}
  \toprule
  \textbf{Dataset} & \textbf{TED-Ed} & \textbf{Khan} & \textbf{Total}\\
  \midrule
  \textbf{\#Train Videos} & \num{1133} & \num{6290} & \num{7423} \\
  \textbf{\#Val Videos} & \num{141} & \num{786} & \num{927} \\
  \textbf{\#Test Videos} & \num{143} & \num{787} & \num{930} \\
  \midrule
  \textbf{\#Train Questions} & \num{5601} & \num{161901} & \num{167502} \\
  \textbf{\#Val Questions} & \num{726} & \num{21970} & \num{22696} \\
  \textbf{\#Test Questions} & \num{680} & \num{20342} & \num{21022} \\
  \bottomrule
\end{tabular}
%}
\end{table}
\subsection{Models}
\label{subsec:models}
Table~\ref{tab:vlm_models} shows the selected models and the characteristics of their input modalities.
Each VLM uses one of two main approaches to frame processing: 
\begin{inparaenum}
\item \emph{Selective frame usage: }\textit{Video-LLaVA}~\cite{lin2023video}, 
\item \emph{Comprehensive frame sampling: }\textit{PG-Video-LLaVA}~\cite{munasinghe2023PGVideoLLaVA}, \textit{Video-LLaMA}~\cite{damonlpsg2023videollama}
\end{inparaenum}
Additionally, we include the instruction-tuned version of the LLM \textit{Mistral-7B}~\cite{mistral_7b} to pair with the VLM model variants since LLMs have already been used for question generation in previous research~\cite{berger2024llms}.
This model takes a prompt and the transcript as input for processing.
Notably, \textit{Video-LLaMA}~\cite{damonlpsg2023videollama} processes the original audio signal, while \textit{PG-Video-LLaVA} transcribes speech. 
We use a zero-shot approach of the monomodal models \textit{Video-LLaVA} and \textit{Mistral-7B} as a baseline to assess the extent to which multimodal models perform.
All models use prompts, allowing analysis of how prompt formulation affects results. 
%\begin{table}[htbp]
\begin{table}[tbp]
\centering
%\fontsize{8}{10}\selectfont
%\def\arraystretch{0.9}
\caption{Selected models and their characteristics for automatic question generation. \textit{All} for \textit{\#Frames} frames extraction at a regular interval. \textit{Audio} describes the use of the original audio signal present, whereas text represents the use of speech transcript.} 
%(e.g. speech transcript).}
\label{tab:vlm_models}
\begin{tabular}{lcccc}
  \toprule
  \textbf{Model} & \textbf{\#Frames} & \textbf{Image} & \textbf{Audio} & \textbf{Text} \\
  \midrule
  Video-LLaVA~\cite{lin2023video} & 8 & \checkmark & \xmark & \xmark\\
  PG-Video-LLaVA~\cite{munasinghe2023PGVideoLLaVA} & All & \checkmark & \xmark & \checkmark\\
  Video-LLaMA~\cite{damonlpsg2023videollama} & All & \checkmark & \checkmark & \xmark\\
  %Valley~\cite{luo2023valley} & All & \checkmark & \xmark & \xmark\\
  %VideoChat~\cite{li2024videochat} & Keyframes & \xmark & \xmark & \checkmark\\
  %VideoChatGPT~\cite{Maaz2023VideoChatGPT} & Keyframes & \xmark & \xmark & \xmark\\
  Mistral-7B~\cite{mistral_7b} & 0 & \xmark & \xmark & \checkmark\\
  \bottomrule
\end{tabular}
\end{table}
\subsection{Prompt Design}
\label{sec:prompt_design}
Prior research has shown that prompt formulation can influence the accuracy, relevance, and coherence of outputs~\cite{chen2023unleashing}.
To investigate this on question generation, we experiment with three distinct prompts:
\begin{enumerate}
    \item Create a question about the video content.
    \item Develop a question that tests comprehension of the video's main idea.
    \item Generate a question to assess the knowledge acquired from the video.
\end{enumerate}
These prompts differ in specificity and focus, addressing different cognitive skills and learning objectives.
This allows models to generate questions at various levels, ranging from simple recall to deeper understanding, particularly concerning  Bloom's taxonomy~\cite{blooms_taxonomy}. 
The first prompt is formulated more generally, allowing a wide range of questions.
The second one focuses on key takeaways and conceptual understanding.
The last prompt targets questions that test the ability to integrate information into a broader context.
We compare questions generated from these prompts to determine the impact of prompt variations on the results.
For all experiments, we use each prompt $n$ times per video (where $n$ is the number of ground-truth questions of a video), changing the formulation from "a question" to "an additional question" in subsequent iterations.

As each model processes its input differently, model-specific considerations must be taken into account:  
\textit{Video-LLaVA} and \textit{Video-LLaMA} use a cache for each prompt to remember already created questions. Preventing the models from generating duplicate questions within the same prompt session.
For \textit{PG-Video-LLaVA} and \textit{Mistral-7B}, we provide a list of already generated questions with each prompt ("The following questions were already generated: \textit{List}") to mimic this behavior. 
For \textit{Mistral-7B}, we include an additional "Transcript: \textit{transcript content}" containing the speech transcript, as this model does not process videos.
\subsection{Metrics}
\label{sec:metrics}
Considering the evaluation of similar tasks~\cite{berger2024llms,ICWSM18LearningQ,survey_metrics}, commonly used metrics include \textit{BLEU}~\cite{bleu}, \textit{ROUGE}~\cite{rouge} and \textit{METEOR}~\cite{meteor}.
Since these  metrics are primarily word-level, we chose \textit{ROUGE-L}\cite{rouge} for word-based evaluation, which  measures the longest common subsequence between generated and ground-truth questions.
It is complemented by \textit{BertScore}\cite{bertscore} to capture semantic similarity beyond exact word matches using contextual embeddings.
We apply the default \textit{roberta-large} model with rescaling for better value distribution. 
For each generated question, we compare it to all ground-truth questions of a video and count the highest score to capture the best match.
Since these metrics rely on similarity to ground-truth questions, they may penalize meaningful questions that differ.
To address this limitation, we introduce our own \textit{Inner-Class-Diff} (\textit{ICD}) metric, which captures aspects not covered by these metrics by comparing generated questions to the respective video's speech transcript (Equation~\ref{eq:icd_formula}). 
\begin{equation}
\label{eq:icd_formula}
\text{ICD}(q, t, \{t_i\}_{i=1}^{N}) = \cos(\textbf{q}, \textbf{t}) - \frac{1}{N}\sum_{i=1}^{N}\cos(\textbf{q}, \textbf{t}_i)
\end{equation}
We use Khan Academy videos with sufficiently specific domain labels, namely \textit{math}, \textit{science}, \textit{computing} or \textit{economics-finance-domain}, excluding TED-Ed videos due to lack of domain labels.
Non-used labels are, e.g., \textit{partner-content} and \textit{resources}.
We compute the cosine similarity of a question embedding \textbf{q} to its video transcript embedding \textbf{t}.
Besides, we average the cosine similarity to transcripts of other videos within the same domain $\mathbf{t}_i$ and subtract this value from the original similarity score. 
We calculate the text embeddings using \textit{all-MiniLM-L6-v2} from \textit{SentenceTransformer}~\cite{sentence_transformer}, averaging sentence embeddings from the transcript, with sentence splitting handled by the \textit{NLTK} sentence tokenizer\cite{nltk}.
The ICD score ranges from \num{-1} to \num{1}, where positive values indicate questions more closely aligned with the original video content, negative values suggest higher similarity to more unrelated content, and a score of 0 implies either domain-overlap across videos or content-unspecific questions.
A limitation is that visually focused questions can not be processed using transcripts, like other text-based metrics.
\section{Comparing VLMs for Video Question Generation}
\label{sec:exp_evaluation}
In this chapter, we evaluate the question-generation capabilities of VLMs across various conditions to identify strengths and weaknesses.
Our evaluation consists of five key components: 
We assess the pre-trained models' ability to generate content-related questions (Section~\ref{subsec:zeroshot}). %, 
Next, we examine the impact of fine-tuning (Section~\ref{subsec:finetune}) and investigate performance with restricted monomodal-modal input (Section~\ref{subsec:ablation}). 
Finally, we analyze the phrasing of the generated questions (Section~\ref{subsec:data_eval}) and conduct a manual evaluation (Section~\ref{subsec:quality}). 

\subsection{Zero-Shot Setting}
\label{subsec:zeroshot}
%\begin{table}[htbp]
\begin{table}[tbp]
\centering
%\fontsize{8}{10}\selectfont
%\def\arraystretch{0.9}
\caption{Zero-shot evaluation of prompts for \textit{Mistral-7B}, \textit{Video-LLaVA}
\textit{PG-Video-LLaVA},  
and \textit{Video-LLaMA}.
Mode describes prompt type (see Section~\ref{sec:prompt_design}), where \textit{Avg.} is the average result. \textit{Quest.}, \textit{Statem.}, and \textit{Empty} denote the percentage of generated questions, statements, and empty outputs, respectively.  Besides, the scores of the metrics \textit{Rouge}, \textit{BertScore} and our metric (ICD) are given.}
\label{tab:zero_shot_questions}
\begin{tabular}{lcrrr|ccc}%{lrrrr|rrr}
  \toprule
  \textbf{Model} & \textbf{Mode} & \textbf{Question} & \textbf{Statement} & \textbf{Empty} & \textbf{Rouge} & \textbf{BertScore} & \textbf{ICD}\\
  \midrule
  % 7B
  %Video-LLaVA~\cite{lin2023video} & 1 & 99.64\,\% & 0.36\,\% & 0.00\,\% \\
  %Video-LLaVA~\cite{lin2023video} & 2 & 100.00\,\% & 0.00\,\% & 0.00\% \\
  %Video-LLaVA~\cite{lin2023video} & 3 & 100.00\,\% & 0.00\,\% & 0.00\,\% \\
  %Video-LLaVA~\cite{lin2023video} & Avg. & 99.88\,\% & 0.12\,\% & 0.00\,\% \\
  %PG-Video-LLaVA~\cite{munasinghe2023PGVideoLLaVA} & 1 & 88.49\,\% & 11.51\% & 0.00\% \\
  %PG-Video-LLaVA~\cite{munasinghe2023PGVideoLLaVA} & 2 & 99.61\,\% & 0.39\,\% & 0.00\,\% \\
  %PG-Video-LLaVA~\cite{munasinghe2023PGVideoLLaVA} & 3 & 99.34\,\% & 0.66\,\% & 0.00\% \\  
  %PG-Video-LLaVA~\cite{munasinghe2023PGVideoLLaVA} & Avg. & 95.81\,\% & 4.19\,\% & 0.00\% \\ 
  Mistral-7B & 1 & 84.18\,\% & 15.82\,\% & \textbf{0.00\,\%} & 0.10  & 0.07 & \textbf{0.37}\\
  Mistral-7B & 2 & 81.22\,\% & 18.78\,\% & \textbf{0.00\,\%} & 0.10 & 0.02 & \textbf{0.37}\\
  Mistral-7B & 3 & 77.88\,\% & 22.12\,\% & \textbf{0.00\,\%} & 0.11 & 0.07 & 0.36\\
  Mistral-7B & Avg. & 81.13\,\% & 18.87\,\% & \textbf{0.00\,\%} & 0.10 & 0.06 & \textbf{0.37}\\

  Video-LLaVA & 1 & \textbf{100.00\,\%} & \textbf{0.00\,\%} & \textbf{0.00\,\%} & \textbf{0.47}  & \textbf{0.18} & 0.06\\
  Video-LLaVA & 2 & 99.99\,\% & 0.01\,\% & \textbf{0.00\,\%} & 0.46 & \textbf{0.18} & 0.04\\
  Video-LLaVA & 3 & \textbf{100.00\,\%} & \textbf{0.00\,\%} & \textbf{0.00\,\%} & 0.43 & 0.16 & 0.10\\
  Video-LLaVA & Avg. & \textbf{100.00\,\%} & \textbf{0.00\,\%} & \textbf{0.00\,\%} & 0.45 & \textbf{0.18} & 0.06\\
  
  PG-Video-LLaVA & 1 & 33.64\,\% & 27.74\% & 38.63\% & 0.10 & 0.02 & 0.07\\
  PG-Video-LLaVA & 2 & 46.63\,\% & 21.92\,\% & 31.45\,\% & 0.11 & 0.02 & 0.08\\
  PG-Video-LLaVA & 3 & 59.81\,\% & 10.35\,\% & 29.84\% & 0.10 & -0.02 & 0.09\\  
  PG-Video-LLaVA & Avg. & 45.87\,\% & 20.50\,\% & 33.63\% & 0.10 & 0.01  & 0.08\\ 

  Video-LLaMA & 1 & 84.49\,\% & 15.51\,\% & \textbf{0.00\,\%} & 0.12 & 0.00 & 0.03\\
  Video-LLaMA & 2 & 92.59\,\% & 7.41\,\% & \textbf{0.00\,\%} & 0.10 & 0.00 & 0.04\\
  Video-LLaMA & 3 & 89.83\,\% & 10.17\,\% & \textbf{0.00\,\%} & 0.12 & 0.02 & 0.04\\
  Video-LLaMA & Avg. & 87.98\,\% & 12.02\,\% & \textbf{0.00\,\%} & 0.12 & 0.01 & 0.04\\
  %Valley~\cite{luo2023valley} & 1 & - & - & - \\
  %Valley~\cite{luo2023valley} & 2 & - & - & - \\
  %Valley~\cite{luo2023valley} & 3 & - & - & - \\
  \bottomrule
\end{tabular}
\end{table}
The zero-shot evaluation of \textit{Mistral-7B}~\cite{mistral_7b} and \textit{Video-LLaVA}~\cite{lin2023video} serves as a baseline comparison for multimodal VLMs.
The performance of the multimodal VLMs is assessed at both the zero-shot and fine-tuned levels (Section~\ref{subsec:finetune}).
Table~\ref{tab:zero_shot_questions} describes different statistics of the models' output.  
On deeper inspection, we observed that \textit{Video-LLaVA} often reformulates the prompt to a question (e.g., "What is the content of the video?").
\textit{PG-Video-LLaVA} implies more problems, producing a mix of unrelated, related questions, summaries, or empty strings. 
\textit{Video-LLaMA} never returns an empty string but sometimes fails to generate a question.
In contrast, the \textit{Mistral-7B} generates more content-related questions. 
When asked for more questions, all models tend to repeat themselves.

For \textit{Rouge} and \textit{BertScore} on the VLM side, \textit{Video-LLaVA} has the highest scores.
Since \textit{Video-LLaVA} only generates short questions, the score automatically increases without providing proper context, as fewer words need to be compared.
The scores of \textit{PG-Video-LLaVA} are the worst since many empty outputs are generated.
Overall, the metrics do not reflect how well the questions match the video content since the question may differ from the ground truth, resulting in a lower score.
This is also visible in the values of \textit{Mistral-7B}, which are not very high despite more content-related generation.
Considering the \textit{ICD} scores, their values are always positive.
This indicates that, on average, the output is more similar to the used video than to others of the same class.
\subsection{Fine-Tuning Setting}
\label{subsec:finetune}
\paragraph{Training approach:}
We fine-tuned \textit{PG-Video-LLaVA} and \textit{Video-LLaMA} on the modified training data of \textit{LearningQ}~\cite{ICWSM18LearningQ} to analyze whether these models can be optimized for multimodal question generation. 
We used the models' code provided by the authors, changed the batch size to two to account for computing capacity limitations, and adjusted the learning rate using the linear scaling rule~\cite{scaling_learning_rate}. 
We use the \textit{13B} versions for both models.
Models are trained on the first prompt (see Section~\ref{subsec:zeroshot}) since it is formulated in a more general way.
\paragraph{Results:} We still used all three prompts to ensure a better comparison with the zero-shot approach.
The first and second entries of Table~\ref{tab:ablation_amount} describe the 
fine-tuned models.
Compared to zero-shot, the number of generated questions increased.
However, \textit{PG-Video-LLaVA} still struggles with Khan Academy videos, likely due to varied handwriting.
This leads to misinterpretations, and combined with transcripts, the model may produce unclear output.
\textit{Video-LLaMA} shows increased stability in generating questions, indicating a more robust adaptation to the task, even with different prompts.
The metrics \textit{ROUGE}, \textit{BERTScore} and \textit{ICD} mostly improved. 
Training on educational videos likely contributed to a better alignment with the task.
Even if no question is generated (see \textit{PG-Video-LLaVA}), the output may be closer to the content, which can lead to improved scores.
This implies manual question analysis is still required for evaluating details like complexity and relevance, and to compare with the baselines.
\begin{table}[htbp]
\centering
%\fontsize{8}{10}\selectfont
%\def\arraystretch{0.9}
\caption{Evaluation of fine-tuned \textit{PG-Video-LLaVA}  and \textit{Video-LLaMA} across modalities (multi-modal, visual only [V], audio only [A]). \emph{Quest.}, \textit{Statem.}, and \textit{Empty} denote the percentage of generated questions, statements, and empty outputs, respectively. Besides, the scores of the metrics \textit{Rouge}, \textit{BertScore} and our metric (ICD) are given.}
\label{tab:ablation_amount}
\begin{tabular}{lrrr|ccc}
  \toprule
  \textbf{Model} & \textbf{Question} & \textbf{Statement} & \textbf{Empty} & \textbf{Rouge} & \textbf{BertScore} & \textbf{ICD}\\
  \midrule
  PG-Video-LLaVA & 52.74\,\% & 12.71\,\% & 34.55\,\% & 0.31 & 0.02 & 0.16\\
  Video-LLaMA & 98.86\,\% & 1.14\,\% & \textbf{0.00\,\%} & 0.22 & 0.11 & 0.11\\
  \midrule
  PG-Video-LLaVA (V) & 97.76\,\% & \textbf{0.15\,\%} & 2.09\,\% & \textbf{0.51} & 0.16 & \textbf{0.23}\\
  PG-Video-LLaVA (A) & 96.66\,\% & 1.24\,\% & 2.10\,\% & 0.49 & \textbf{0.18} & 0.03\\
  Video-LLaMA (V) & 72.68\,\% & 24.12\,\% & 3.21\,\% & 0.18 & 0.12 & 0.07\\
  Video-LLaMA (A) & \textbf{99.33\,\%} & 0.67\,\% & \textbf{0.00\,\%} & 0.21 & 0.10 & 0.00\\
  \bottomrule
\end{tabular}
\end{table}
\subsection{Ablation Study}
\label{subsec:ablation}
Given the multimodal nature of videos, it is unclear how important each modality is.
To gain a better insight, we evaluate the fine-tuned models with only one modality: replacing frames with black frames in one experiment and removing the audio track in another (second part of Table~\ref{tab:ablation_amount}).

\textit{PG-Video-LLaVA} mostly generates questions using only frames or transcripts.
The outputs revealed that the model hallucinates when lacking one modality, generating more non-content-specific questions.
In contrast, \textit{Video-LLaMA} generates 
fewer questions when using only frames, likely due to 
handwriting quality in Khan Academy videos, while the audio-only approach performs similarly to the multimodal method.

Focusing on the metrics, the results for the \textit{PG-Video-LLaVA} overall improve, likely due to more generated short questions, since they are more similar to the ground-truth questions structure, e.g., using the same question word.
Still, the \textit{ICD} value of the audio-only approach is getting worse, given the impression that there is hardly any similarity to the content.
For \textit{Video-LLaMA}, almost all metric values are worse, which is caused by the fewer content-specific questions.
For a more detailed analysis, datasets with more diverse video presentation styles are required to clarify each modality's significance within each style.
\subsection{Evaluation of Question Structure}
\label{subsec:data_eval}
\begin{table*}[tb]
\centering
%\fontsize{8}{10}\selectfont
%\def\arraystretch{0.9}
\caption{Distribution of question words (\%) of \textit{Mistral-7B} (M-7B),  \textit{Video-LLaVA} (V-L), \textit{PG-Video-LLaVA} (P-V-L) and \textit{Video-LLaMA} (V-LM) on generated questions across approaches (GT: ground truth. *: fine-tuned; \textit{V}: frames only; \textit{A}: audio only). \textit{None} defines questions without a question word.}
\label{tab:zeroshot_questionwords}
\begin{tabular}{lrrrrrrrrrr}
  \toprule
  \textbf{Model} & \textbf{Where} & \textbf{Who} & \textbf{When} & \textbf{What} & \textbf{Why} & \textbf{Whose}
   & \textbf{Whom} & \textbf{Which} & \textbf{How} & \textbf{None}\\
  \midrule
  M-7B & 16.42 & 2.17 & 14.89 & 44.37 & 4.17 & 0.09 & 0.00 & 4.80 & 11.08 & 2.01\\
  V-L & 0.21 & 2.20 & 1.63 & 73.73 & 0.00 & 0.00 & 0.00 & 0.00 & 21.97 & 0.26 \\
  P-V-L & 4.37 & 2.76 & 3.81 & 43.20 & 0.17 & 0.00 & 0.00 &10.54 & 20.51 & 14.65 \\
  V-LM & 3.04 & 1.49 & 1.31 & 41.91 & 0.59 & 0.00 & 0.00 & 5.32 & 25.11 & 21.24 \\
  \midrule
  P-V-L* & 0.20 & 0.52 & 3.09 & 72.98 & 6.84 & 0.00 & 0.00 & 0.15 & 14.42 & 1.80 \\
  P-V-L* (V) & 0.24 & 0.27 & 4.68 & 44.95 & 18.51 & 0.00 & 0.00 & 2.04 & 25.59 & 3.73 \\
  P-V-L* (A) & 0.05 & 0.52 & 0.10 & 97.76 & 0.03 & 0.00 & 0.00 & 0.01 & 0.66  & 0.86\\
  
  V-LM* & 2.41 & 0.97 & 3.72 & 42.56 & 3.91 & 0.01 & 0.00 & 2.31  & 28.47 & 15.63 \\
  V-LM* (V) & 1.92 & 1.07 & 4.10 & 45.67 & 3.22 & 0.01 & 0.00 & 2.31 & 30.86 & 10.84 \\
  V-LM* (A) & 1.58 & 1.17 & 3.43 & 40.95 & 3.22 & 0.00 & 0.00 & 3.30 & 32.92 & 13.43 \\
  \midrule
  GT & 2.62 & 0.88 & 7.67 & 19.90 & 13.58 & 0.05 & 0.01 & 2.25 & 17.29 & 35.76\\
  \bottomrule
\end{tabular}
\end{table*}
We analyze question words generated by different models, averaging over all three prompts and considering \textit{wh-words} and \textit{how}.
Besides, we analyze sentence length (minimum, average, maximum) and readability (\textit{Flesch's index}~\cite{flesch}).
Like \textit{LearningQ}~\cite{ICWSM18LearningQ}, we use Flesch to compare readability across data.
This analysis aims to compare characteristics between generated questions and ground truth.

Table~\ref{tab:zeroshot_questionwords} describes the percentage distribution of question words.
All models predominantly generate questions using \textit{what}.
For fine-tuning, most ground-truth questions lack explicit question words, and when present, \textit{what} is the most frequently used, influencing the learning process, and leading models to generate more "what" questions.
Zero-shot models' behavior suggests that they learned this focus on tasks like question answering.
\textit{PG-Video-LLaVA} shows significant variation, likely due to unpredictable behavior.
For \textit{Video-LLaMA}, the distribution of all approaches is consistent.
In contrast, \textit{Mistral-7B} has a broader variety, suggesting a brighter understanding.
Overall, a more balanced fine-tuning dataset might lead to a more varied distribution.

Table~\ref{tab:zeroshot_stats} reports the results 
on output length and readability. 
For zero-shot, \textit{PG-Video-LLaVA} and \textit{Video-LLaMA} generate significantly longer sentences on average than \textit{Video-LLaVA}.
This also applies for \textit{Mistral-7B}, where generated questions often include answers.
The models' readability indicates that the text appears harder to read than the ground truth, while \textit{Video-LLaVA} produces more readable outputs.
Fine-tuning improves readability and shortens outputs, as the training data consists of shorter questions.
Furthermore, a single word with "?" can serve as a question, which is reflected in fine-tuned models.
\begin{table}[tb]
\centering
%\fontsize{8}{10}\selectfont
%\def\arraystretch{0.9}
\caption{Statistics of \textit{Mistral-7B}, \textit{Video-LLaVA} %~\cite{lin2023video}
, \textit{PG-Video-LLaVA} %~\cite{munasinghe2023PGVideoLLaVA} 
and \textit{Video-LLaMA} %~\cite{damonlpsg2023videollama} 
output across approaches (GT: ground truth. *: fine-tuned. \textit{V}: frames only. \textit{A}: audio only).
Text length: \textit{Min} stands for minimum, \textit{Avg.} for average and \textit{Max} for maximum. \textit{Flesch} represents the score of Flesch's index.}
\label{tab:zeroshot_stats}
\begin{tabular}{lrrrr}
  \toprule
  \textbf{Model} & \textbf{Min} & \textbf{Avg.} & \textbf{Max} & \textbf{Flesch}\\
  \midrule
  Mistral-7B & 11 & 148.03 & 275 & 41.30 \\
  Video-LLaVA & 5 & 11.64 & 50 & 86.14 \\
  PG-Video-LLaVA & 4 & 75.68 & 710 & 56.47 \\
  Video-LLaMA & 7 & 70.6 & 267 & 55.39 \\
  \midrule
  PG-Video-LLaVA* & 1 & 11.98 & 824 & 73.35 \\
  PG-Video-LLaVA* (V) & 1 & 10.89 & 54 & 77.41 \\
  PG-Video-LLaVA* (A) & 3 & 9.69 & 783 & 82.33 \\
  Video-LLaMA* & 3 & 28.64 & 257 & 62.06 \\
  Video-LLaMA* (V) & 1 & 32.60 & 285 & 58.64 \\
  Video-LLaMA* (A) & 3 & 29.55 & 265 & 62.89 \\
  \midrule
  GT & 1 & 14.70 & 244 & 73.02 \\
  \bottomrule
\end{tabular}
\end{table}
\subsection{Qualitative Evaluation}
\label{subsec:quality}
Since most automatic evaluation metrics rely on ground-truth questions, they may not accurately assess whether a generated question is meaningful if it differs. 
To investigate this, we qualitatively evaluate questions from six videos (three TED-ED and three Khan Academy videos) across three dimensions:
\begin{itemize}
    \item \textbf{Relevance:} Does the generated question refer to the video content?
    \item \textbf{Answerability:} Can the question be answered based on the video content?
    \item  \textbf{Levels of understanding:} Does the generated question require a certain level of understanding from the learner w.r.t. Bloom's taxonomy?
\end{itemize}
Relevance and answerability (binary values) are used to investigate if a generated question is meaningful and grounded in the video content similar to Elkins et al.~\cite{relevance_edu_qg}.
Bloom's taxonomy~\cite{blooms_taxonomy} was chosen for levels of understanding, as in \textit{LearningQ}~\cite{ICWSM18LearningQ} and related research~\cite{qg_bloom2,qg_school_level,qg_bloom}.
For each approach (zero-shot and fine-tuned), two of the authors independently evaluated the first generated response from each prompt, followed by a discussion.  
Additionally, Krippendorff's alpha is calculated to assess the inter-rater reliability before discussion.
A total of $180$ outputs is analyzed, with results and code available online\footnote{\url{https://github.com/markossta/aied_2025_video_qg}}.

Krippendorff's alpha scores indicate moderate to good agreement for relevance and answerability (\num{0.71}, \num{0.79}).
In contrast, levels of understanding have a low agreement (\num{0.43}), reflecting the complexity of evaluating levels of understanding.
This confirms inconsistencies, indicating the need for a more collaborative approach.
\begin{table*}[tb]
\centering
%\fontsize{8}{10}\selectfont
%\def\arraystretch{0.9}
\caption{Results of manual evaluation (\%) for \textit{Mistral-7B} (M-7B), \textit{Video-LLaVA} (V-L), \textit{PG-Video-LLaVA} (P-V-L) and \textit{Video-LLaMA} (V-LM) across approaches (*: fine-tuned; \textit{V}: frames only; \textit{A}: audio only). The evaluation metrics are defined as \textit{Rel.}: relevance, \textit{Ans.}: answerability. and understanding levels: \textit{non} (no level), \textit{rem.} (remember), \textit{underst.} (understand), \textit{apply}, \textit{analyze}, \textit{eval.} (evaluate) and \textit{create}.
}
\label{tab:percentage_blooms}
\begin{tabular}{lrr|rrrrrrr}
  \toprule
  \textbf{Model} & \textbf{Rel.} & \textbf{Ans.} & \textbf{Non} & \textbf{Rem.} & \textbf{Underst.} & \textbf{Apply} & \textbf{Analyze} & \textbf{Eval.} & \textbf{Create}\\
  \midrule
  M-7B & \textbf{100.00} & \textbf{93.33} & 0.00 & 33.33 & 33.33 & \textbf{13.33} & 0.00 & 20.00 & 0.00\\
  V-L & 88.24 & 88.24 & 0.00 & 23.53 & 70.59 & 0.00 & 5.88 & 0.00 & 0.00\\
  P-V-L & 70.59 & 70.59 & 0.00 & 35.29 & 47.06 & 0.00 & 5.88 & 11.76 & 0.00\\
  V-LM & 75.00 & 25.00 & \textbf{50.00} & 0.00 & 25.00 & 0.00 & 0.00 & 0.00 & \textbf{25.00}\\
  \midrule
  P-V-L* & 93.75 & 87.50 & 0.00 & 43.75 & \textbf{50.00} & 0.00 & 0.00 & 6.25 & 0.00\\
  P-V-L* (V) & 61.11 & 38.89 & 0.00 & \textbf{55.56} & 27.78 & 5.56 & 0.00 & 11.11 & 0.00\\
  P-V-L* (A) & 73.33 & 73.33 & 0.00 & 46.67 & 40.00 & 0.00 & 13.33 & 0.00 & 0.00\\
  
  V-LM* & 23.53 & 5.88 & 23.53 & 11.76 & 23.53 & 0.00 & \textbf{17.65} & \textbf{23.53} & 0.00\\
  V-LM* (V) & 50.00 & 11.11 & 11.11 & 38.89 & 16.67 & 0.00 & 16.67 & 16.67 & 0.00\\
  V-LM* (A) & 0.00 & 0.00 & 22.22 & 33.33 & 22.22 & 0.00 & 16.67 & 5.56 & 0.00\\
  \midrule
  Overall & 61.29 & 49.68 & 7.74 & 34.84 & 36.13 & 1.94 & 8.39 & 10.32 & 0.65\\
  \bottomrule
\end{tabular}
\end{table*}
Table~\ref{tab:percentage_blooms} shows the results after discussion.
We filtered out non-question outputs to focus on the question structure.
looking at \textit{relevance} and \textit{answerability}, the baselines (\textit{Mistral-7B}, \textit{Video-LLaVA}) generate mostly relevant output.
\textit{Mistral-7B} consistently produces answerable questions, often providing answer options.
This suggests that it has already a more effective task performance without fine-tuning.
\textit{Video-LLaVA} also performs well since we consider questions like "What is the main idea of the video?" to be content-relevant and answerable.
However, these questions only assess a superficial understanding of the video. 
The model did not generate questions about specific parts of the video.
Looking at the VLMs, it is evident that, for \textit{PG-Video-LLaVA}, both metrics increased from zero-shot to fine-tuned, indicating improved task adaptation.
The values decreased for \textit{Video-LLaMA}, but this is due to the low number of generated questions in zero-shot (mostly statements as output), resulting in little variation.
It is the only model that generated questions outside of Bloom's taxonomy, like the non-knowledge-focused question: "What would you like me to ask about these illustrations?".
Focusing on audio-only and frames-only approaches, \textit{PG-Video-LLaVA} performs worse for both but still generates valid results, with questions sometimes lacking depth.
In contrast, \textit{Video-LLaMA} improves when using only frames, indicating both models can create questions without transcripts/audio.
Considering the levels of understanding, the approaches show a similar distribution, focusing mainly on \textit{remembering} and \textit{understanding}. 
However, the frequency of \textit{analyzing} and \textit{evaluating} varies by model.
Overall, the multimodal VLMs require a better combination of audio and visual features to produce a more accurate output like \textit{Mistral-7B}. 

\section{Conclusions}
\label{sec:conclusion}
In this paper, we have investigated the capabilities of state-of-the-art VLMs to generate questions for educational videos. 
Our key findings include: 
\begin{inparaenum}[(1)]
    \item  Zero-shot approaches struggled with the task, generating superficial or irrelevant questions due to a lack of exposure to educational content during training (RQ1).
    \item Fine-tuning improved performance, but some generated questions remained content-irrelevant, indicating a need for more sophisticated fine-tuning techniques and larger, more diverse datasets or alternative approaches (RQ2).
    The textual baseline (\textit{Mistral-7B}) not using frames results in better performance, as it can focus on the text modality, highlighting the need for improved methods to process multimodal information.
    \item The importance of individual modalities (e.g., visual, audio) varies across %different 
    educational video types (e.g., slides, animations, handwriting), necessitating research to optimize models for each format (RQ4).
    \item The study revealed significant gaps in available datasets. While resources like LearningQ~\cite{ICWSM18LearningQ} provide \num{9000} videos, they fail to encompass the full spectrum of educational video types~\cite{video_types} (RQ3). 
    Moreover, existing datasets lack time-stamp information, making it challenging to locate relevant video segments (RQ4). 
\end{inparaenum}

In Future, 
developing larger, more diverse datasets that better represent various educational video styles and include time-stamped questions for precise content-question alignment is crucial. 
Further research should focus on conducting in-depth analyses of different video representation styles to assess their impact on models. 
Exploring advanced fine-tuning techniques could improve the content relevance of generated questions.
Additionally, future work could involve refining prompt analysis techniques to gain deeper insights into their influence on various aspects.
Finally, alternative approaches like ensemble methods should be investigated, including non-generative models, e.g., for analyzing visual content or avoiding hallucinations, and combining them with LLM or VLM advantages.

\begin{credits}
\subsubsection{\ackname} This work has been financially supported by the Lower Saxony Ministry of Science and Culture with funds from the zukunft.niedersachsen program of the VolkswagenStiftung for the project "VidQA: Automated comprehension tests for learning videos".

\subsubsection{\discintname}
The authors have no competing interests to declare that are relevant to the content of this article.
\end{credits}

\bibliographystyle{splncs04}
\bibliography{bibliography}

\begin{thebibliography}{10}
\providecommand{\url}[1]{\texttt{#1}}
\providecommand{\urlprefix}{URL }
\providecommand{\doi}[1]{https://doi.org/#1}

\bibitem{qg_graessers}
{Al Faraby}, S., Romadhony, A., Adiwijaya: Analysis of llms for educational question classification and generation. Computers and Education: Artificial Intelligence  \textbf{7},  100298 (2024). \doi{10.1016/j.caeai.2024.100298}

\bibitem{awadalla2023openflamingo}
Awadalla, A., Gao, I., Gardner, J., Hessel, J., Hanafy, Y., Zhu, W., Marathe, K., Bitton, Y., Gadre, S.Y., Sagawa, S., Jitsev, J., Kornblith, S., Koh, P.W., Ilharco, G., Wortsman, M., Schmidt, L.: Openflamingo: An open-source framework for training large autoregressive vision-language models. arXiv preprint  \textbf{abs/2308.01390} (2023). \doi{10.48550/ARXIV.2308.01390}

\bibitem{berger2024llms}
Berger, J., Koß, J., Stamatakis, M., Hoppe, A., Ewerth, R., Wartena, C.: Question generation capabilities of ``small`` large language models. In: International Conference on Natural Language \& Information Systems, {NLDB} 2024, Turin, Italy, June 25-27, 2024. pp. 183--194. Springer (2024). \doi{10.1007/978-3-031-70242-6\_18}

\bibitem{relevant_video_qg}
Bhanot, G., Pal, B., Sah, B., Chhikara, G.: Generating relevant question answer from video. In: Conference on Computing Communication and Networking Technologies, {ICCCNT} 2023, Delhi, India, July 6-8, 2023. pp.~1--6. {IEEE} (2023). \doi{10.1109/ICCCNT56998.2023.10306677}

\bibitem{mcq_educator}
Biancini, G., Ferrato, A., Limongelli, C.: Multiple-choice question generation using large language models: Methodology and educator insights. In: Conference on User Modeling, Adaptation and Personalization, {UMAP} 2024, Cagliari, Italy, July 1-4, 2024. {ACM} (2024). \doi{10.1145/3631700.3665233}

\bibitem{nltk}
Bird, S., Klein, E., Loper, E.: Natural language processing with Python: analyzing text with the natural language toolkit. " O'Reilly Media, Inc." (2009)

\bibitem{graessers_taxonomy}
Cao, S., Wang, L.: Controllable open-ended question generation with {A} new question type ontology. In: Annual Meeting of the Association for Computational Linguistics and the 11th International Joint Conference on Natural Language Processing, {ACL/IJCNLP} 2021, (Volume 1: Long Papers), Virtual Event, August 1-6, 2021. pp. 6424--6439. Association for Computational Linguistics (2021). \doi{10.18653/V1/2021.ACL-LONG.502}

\bibitem{chen2023unleashing}
Chen, B., Zhang, Z., Langren{\'{e}}, N., Zhu, S.: Unleashing the potential of prompt engineering in large language models: a comprehensive review. arXiv preprint  \textbf{abs/2310.14735} (2023). \doi{10.48550/ARXIV.2310.14735}

\bibitem{ICWSM18LearningQ}
Chen, G., Yang, J., Hauff, C., Houben, G.: Learningq: {A} large-scale dataset for educational question generation. In: International Conference on Web and Social Media, {ICWSM} 2018, Stanford, (California), USA, June 25-28, 2018. pp. 481--490. {AAAI} Press (2018). \doi{10.1609/icwsm.v12i1.14987}

\bibitem{tutorialvqa}
Colas, A.M., Kim, S., Dernoncourt, F., Gupte, S., Wang, D.Z., Kim, D.S.: Tutorialvqa: Question answering dataset for tutorial videos. In: Language Resources and Evaluation Conference, {LREC} 2020, Marseille, France, May 11-16, 2020. pp. 5450--5455 (2020), \url{https://aclanthology.org/2020.lrec-1.670/}

\bibitem{blooms_taxonomy}
Conklin, J.: A taxonomy for learning, teaching, and assessing: A revision of bloom's taxonomy of educational objectives complete edition. Educational Horizons  \textbf{83}(3),  154--159 (2005), \url{http://www.jstor.org/stable/42926529}

\bibitem{relevance_edu_qg}
Elkins, S., Kochmar, E., Serban, I., Cheung, J.C.K.: How useful are educational questions generated by large language models? In: Artificial Intelligence in Education, {AIED} 2023, Tokyo, Japan, July 3-7, 2023. pp. 536--542. Springer (2023). \doi{10.1007/978-3-031-36336-8\_83}

\bibitem{flesch}
Flesch, R.: A new readability yardstick. Journal of applied psychology  \textbf{32}(3), ~221 (1948). \doi{10.1037/h0057532}

\bibitem{videodl}
Forkan, A.R.M., Kang, Y., Jayaraman, P.P., Du, H., Thomson, S., Kollias, E., Wieland, N.: Videodl: Video-based digital learning framework using {AI} question generation and answer assessment. International Journal of Advanced Corporate Learning  \textbf{16}(1),  19--27 (2023). \doi{10.3991/IJAC.V16I1.35207}

\bibitem{scaling_learning_rate}
Goyal, P., Doll{\'{a}}r, P., Girshick, R.B., Noordhuis, P., Wesolowski, L., Kyrola, A., Tulloch, A., Jia, Y., He, K.: Accurate, large minibatch {SGD:} training imagenet in 1 hour. arXiv preprint  \textbf{abs/1706.02677} (2017), \url{http://arxiv.org/abs/1706.02677}

\bibitem{engagment_learning}
da~Gra{\c{c}}a Campos~Pimentel, M., Yaguinuma, C.A., Martins, D.S., Zaine, I.: Anchoring interactive points of interest on web-based instructional video: effects on students' interaction behavior and perceived experience. In: {ACM/SIGAPP} Symposium on Applied Computing, {SAC} 2019, Limassol, Cyprus, April 8-12, 2019. pp. 2445--2452. {ACM} (2019). \doi{10.1145/3297280.3297521}

\bibitem{gupta2023dataset}
Gupta, D., Attal, K., Demner-Fushman, D.: A dataset for medical instructional video classification and question answering. Scientific Data  \textbf{10}(1), ~158 (2023). \doi{10.1038/s41597-023-02036-y}

\bibitem{eduqg}
Hadifar, A., Bitew, S.K., Deleu, J., Develder, C., Demeester, T.: Eduqg: {A} multi-format multiple-choice dataset for the educational domain. {IEEE} Access  \textbf{11},  20885--20896 (2023). \doi{10.1109/ACCESS.2023.3248790}

\bibitem{qg_bloom2}
Hwang, K., Wang, K., Alomair, M., Choa, F., Chen, L.K.: Towards automated multiple choice question generation and evaluation: Aligning with bloom's taxonomy. In: Artificial Intelligence in Education, {AIED} 2024, Recife, Brazil, July 8-12, 2024. pp. 389--396. Springer (2024). \doi{10.1007/978-3-031-64299-9\_35}

\bibitem{mistral_7b}
Jiang, A.Q., Sablayrolles, A., Mensch, A., Bamford, C., Chaplot, D.S., de~Las~Casas, D., Bressand, F., Lengyel, G., Lample, G., Saulnier, L., Lavaud, L.R., Lachaux, M., Stock, P., Scao, T.L., Lavril, T., Wang, T., Lacroix, T., Sayed, W.E.: Mistral 7b. arXiv preprint  \textbf{abs/2310.06825} (2023). \doi{10.48550/ARXIV.2310.06825}

\bibitem{tqa}
Kembhavi, A., Seo, M.J., Schwenk, D., Choi, J., Farhadi, A., Hajishirzi, H.: Are you smarter than a sixth grader? textbook question answering for multimodal machine comprehension. In: {IEEE} Conference on Computer Vision and Pattern Recognition, {CVPR} 2017, Honolulu, (Hawaii), USA, July 21-26, 2017. pp. 5376--5384. {IEEE} Computer Society (2017). \doi{10.1109/CVPR.2017.571}

\bibitem{meteor}
Lavie, A., Agarwal, A.: {METEOR:} an automatic metric for {MT} evaluation with high levels of correlation with human judgments. In: Workshop on Statistical Machine Translation co-locared with the ACL Conference, WMT@ACL 2007, Prague, Czech Republic, June 23, 2007. pp. 228--231. Association for Computational Linguistics (2007), \url{https://aclanthology.org/W07-0734/}

\bibitem{lei2018tvqa}
Lei, J., Yu, L., Bansal, M., Berg, T.L.: {TVQA:} localized, compositional video question answering. In: Proceedings of the 2018 Conference on Empirical Methods in Natural Language Processing, Brussels, Belgium, October 31 - November 4, 2018. pp. 1369--1379. Association for Computational Linguistics (2018). \doi{10.18653/V1/D18-1167}

\bibitem{lei2019tvqa}
Lei, J., Yu, L., Berg, T.L., Bansal, M.: {TVQA+:} spatio-temporal grounding for video question answering. In: Association for Computational Linguistics, {ACL} 2020, Virtual Event, July 5-10, 2020. pp. 8211--8225. Association for Computational Linguistics (2020). \doi{10.18653/V1/2020.ACL-MAIN.730}

\bibitem{Leisner2020DifferentWO}
Leisner, D., Zahn, C.G., Ruf, A., Cattaneo, A.A.P.: Different ways of interacting with videos during learning in secondary physics lessons. In: {HCI} International Conference, {HCII} 2020, Copenhagen, Denmark, July 19-24, 2020. pp. 284--291. Springer (2020). \doi{10.1007/978-3-030-50729-9\_40}

\bibitem{Li2023BLIP2BL}
Li, J., Li, D., Savarese, S., Hoi, S.C.H.: {BLIP-2:} bootstrapping language-image pre-training with frozen image encoders and large language models. In: International Conference on Machine Learning, {ICML} 2023, Honolulu, (Hawaii), USA, July 23-29 2023. pp. 19730--19742. {PMLR} (2023), \url{https://proceedings.mlr.press/v202/li23q.html}

\bibitem{li2024videochat}
Li, K., He, Y., Wang, Y., Li, Y., Wang, W., Luo, P., Wang, Y., Wang, L., Qiao, Y.: Videochat: Chat-centric video understanding. arXiv preprint  \textbf{abs/2305.06355} (2023). \doi{10.48550/ARXIV.2305.06355}

\bibitem{topic_controlled_qg}
Li, Z., Cukurova, M., Bulathwela, S.: A novel approach to scalable and automatic topic-controlled question generation in education (2025). \doi{10.48550/ARXIV.2501.05220}

\bibitem{lin2023video}
Lin, B., Ye, Y., Zhu, B., Cui, J., Ning, M., Jin, P., Yuan, L.: Video-llava: Learning united visual representation by alignment before projection. arXiv preprint  \textbf{abs/2311.10122} (2023). \doi{10.48550/ARXIV.2311.10122}

\bibitem{rouge}
Lin, C.Y.: {ROUGE}: A package for automatic evaluation of summaries. In: Association for Computational Linguistics, {ACL} 2004, Barcelona, Spain, July 21-26, 2004. pp. 74--81. Association for Computational Linguistics (2004), \url{https://aclanthology.org/W04-1013}

\bibitem{liu2023improvedllava}
Liu, H., Li, C., Li, Y., Lee, Y.J.: Improved baselines with visual instruction tuning. arXiv preprint  \textbf{abs/2310.03744} (2023). \doi{10.48550/ARXIV.2310.03744}

\bibitem{liu2023llava}
Liu, H., Li, C., Wu, Q., Lee, Y.J.: Visual instruction tuning. In: Advances in Neural Information Processing Systems, {NeurIPS} 2023, New Orleans, (LA), USA, December 10 - 16, 2023. vol.~36, pp. 34892--34916. Curran Associates, Inc. (2023), \url{https://proceedings.neurips.cc/paper\_files/paper/2023/file/6dcf277ea32ce3288914faf369fe6de0-Paper-Conference.pdf}

\bibitem{luo2023valley}
Luo, R., Zhao, Z., Yang, M., Dong, J., Qiu, M., Lu, P., Wang, T., Wei, Z.: Valley: Video assistant with large language model enhanced ability. arXiv preprint  \textbf{abs/2306.07207} (2023). \doi{10.48550/ARXIV.2306.07207}

\bibitem{Maaz2023VideoChatGPT}
Maaz, M., Rasheed, H.A., Khan, S.H., Khan, F.S.: Video-chatgpt: Towards detailed video understanding via large vision and language models. arXiv preprint  \textbf{abs/2306.05424} (2023). \doi{10.48550/ARXIV.2306.05424}

\bibitem{qg_school_level}
Maity, S., Deroy, A., Sarkar, S.: Can large language models meet the challenge of generating school-level questions? Computers and Education: Artificial Intelligence  \textbf{8},  100370 (2025). \doi{10.1016/j.caeai.2025.100370}

\bibitem{MERKT2022104355}
Merkt, M., Hoppe, A., Bruns, G., Ewerth, R., Huff, M.: Pushing the button: Why do learners pause online videos? Computers \& Education  \textbf{176},  104355 (2022). \doi{10.1016/j.compedu.2021.104355}

\bibitem{qg_university}
Mucciaccia, S.S., Paix{\~{a}}o, T.M., Mutz, F.W., Badue, C.S., de~Souza, A.F., Oliveira{-}Santos, T.: Automatic multiple-choice question generation and evaluation systems based on {LLM:} {A} study case with university resolutions. In: Conference on Computational Linguistics, {COLING} 2025, Abu Dhabi, UAE, January 19-24, 2025. pp. 2246--2260. Association for Computational Linguistics (2025), \url{https://aclanthology.org/2025.coling-main.154/}

\bibitem{munasinghe2023PGVideoLLaVA}
Munasinghe, S., Thushara, R., Maaz, M., Rasheed, H.A., Khan, S., Shah, M., Khan, F.S.: Pg-video-llava: Pixel grounding large video-language models. arXiv preprint  \textbf{abs/2311.13435} (2023). \doi{10.48550/ARXIV.2311.13435}

\bibitem{openai2023gpt4}
OpenAI: {GPT-4} technical report. arXiv preprint  \textbf{abs/2303.08774} (2023). \doi{10.48550/ARXIV.2303.08774}

\bibitem{self_paced_learning}
Palaigeorgiou, G., Papadopoulou, A.: Promoting self-paced learning in the elementary classroom with interactive video, an online course platform and tablets. Education and Information Technologies  \textbf{24}(1),  805--823 (2019). \doi{10.1007/S10639-018-9804-5}

\bibitem{bleu}
Papineni, K., Roukos, S., Ward, T., Zhu, W.: Bleu: a method for automatic evaluation of machine translation. In: Association for Computational Linguistics, {ACL} 2002, Philadelphia, (PA), USA, July 6-12, 2002. pp. 311--318. {ACL} (2002). \doi{10.3115/1073083.1073135}

\bibitem{rajpurkar2016squad}
Rajpurkar, P., Zhang, J., Lopyrev, K., Liang, P.: Squad: 100,000+ questions for machine comprehension of text. In: Conference on Empirical Methods in Natural Language Processing, {EMNLP} 2016, Austin, Texas, USA. pp. 2383--2392. The Association for Computational Linguistics (2016). \doi{10.18653/v1/D16-1264}

\bibitem{sentence_transformer}
Reimers, N., Gurevych, I.: Sentence-bert: Sentence embeddings using siamese bert-networks. In: Conference on Empirical Methods in Natural Language Processing, {EMNLP} 2019, Hong Kong, China, November 3-7, 2019. pp. 3980--3990. Association for Computational Linguistics (2019). \doi{10.18653/V1/D19-1410}

\bibitem{survey_metrics}
Sai, A.B., Mohankumar, A.K., Khapra, M.M.: A survey of evaluation metrics used for {NLG} systems. {ACM} Computing Surveys  \textbf{55}(2),  26:1--26:39 (2023). \doi{10.1145/3485766}

\bibitem{qg_bloom}
Scaria, N., Chenna, S.D., Subramani, D.N.: Automated educational question generation at different bloom's skill levels using large language models: Strategies and evaluation. In: Artificial Intelligence in Education, {AIED} 2024, Recife, Brazil, July 8-12, 2024. pp. 165--179. Springer (2024). \doi{10.1007/978-3-031-64299-9\_12}

\bibitem{survey_nlp}
Shahin, N., Ismail, L.: From rule-based models to deep learning transformers architectures for natural language processing and sign language translation systems: survey, taxonomy and performance evaluation. Artificial Intelligence Review  \textbf{57}(10), ~271 (2024). \doi{10.1007/S10462-024-10895-Z}

\bibitem{DBLP:conf/aied/ShimmeiBM23}
Shimmei, M., Bier, N.L., Matsuda, N.: Machine-generated questions attract instructors when acquainted with learning objectives. In: Artificial Intelligence in Education, {AIED} 2023, Tokyo, Japan, July 3-7, 2023, Proceedings. pp. 3--15. Springer (2023). \doi{10.1007/978-3-031-36272-9\_1}

\bibitem{video_types}
Shoufan, A., Mohamed, F.: On the likes and dislikes of youtube's educational videos: A quantitative study. In: Annual Conference on Information Technology Education, {SIGITE} 2017, New York, USA, October 4 - 7, 2017. p. 127–132. Association for Computing Machinery (2017). \doi{10.1145/3125659.3125692}

\bibitem{skanda_qg}
Skanda, V.C., Jayaram, R., Bukitagar, V.C., Kumar, N.S.: Automatic questionnaire and interactive session generation from videos. In: Computational Intelligence in Data Science, {ICCIDS} 2020, Chennai, India, February 20-22, 2020. pp. 205--212. Springer (2020). \doi{10.1007/978-3-030-63467-4\_16}

\bibitem{xu2017video}
Xu, D., Zhao, Z., Xiao, J., Wu, F., Zhang, H., He, X., Zhuang, Y.: Video question answering via gradually refined attention over appearance and motion. In: {ACM} on Multimedia Conference, {MM} 2017, Mountain View, (CA), USA, October 23-27, 2017. pp. 1645--1653. {ACM} (2017). \doi{10.1145/3123266.3123427}

\bibitem{wikiQA}
Yang, Y., Yih, W., Meek, C.: Wikiqa: {A} challenge dataset for open-domain question answering. In: M{\`{a}}rquez, L., Callison{-}Burch, C., Su, J., Pighin, D., Marton, Y. (eds.) Conference on Empirical Methods in Natural Language Processing, {EMNLP} 2015, Lisbon, Portugal, September 17-21, 2015. pp. 2013--2018. The Association for Computational Linguistics (2015). \doi{10.18653/V1/D15-1237}

\bibitem{medicalexam_qg}
Yao, Z., Parashar, A., Zhou, H., Jang, W.S., Ouyang, F., Yang, Z., Yu, H.: Mcqg-srefine: Multiple choice question generation and evaluation with iterative self-critique, correction, and comparison feedback. arXiv preprint  \textbf{abs/2410.13191} (2024). \doi{10.48550/ARXIV.2410.13191}

\bibitem{yu2019activityqa}
Yu, Z., Xu, D., Yu, J., Yu, T., Zhao, Z., Zhuang, Y., Tao, D.: Activitynet-qa: {A} dataset for understanding complex web videos via question answering. In: The Thirty-Third {AAAI} Conference on Artificial Intelligence, {AAAI} 2019, The Thirty-First Innovative Applications of Artificial Intelligence Conference, {IAAI} 2019, The Ninth {AAAI} Symposium on Educational Advances in Artificial Intelligence, {EAAI} 2019, Honolulu, (Hawaii), USA, January 27 - February 1, 2019. pp. 9127--9134. {AAAI} Press (2019). \doi{10.1609/AAAI.V33I01.33019127}

\bibitem{damonlpsg2023videollama}
Zhang, H., Li, X., Bing, L.: Video-llama: An instruction-tuned audio-visual language model for video understanding. In: Conference on Empirical Methods in Natural Language Processing - System Demonstrations, {EMNLP} 2023, Singapore, December 6-10, 2023. pp. 543--553. Association for Computational Linguistics (2023). \doi{10.18653/V1/2023.EMNLP-DEMO.49}

\bibitem{bertscore}
Zhang, T., Kishore, V., Wu, F., Weinberger, K.Q., Artzi, Y.: Bertscore: Evaluating text generation with {BERT}. In: International Conference on Learning Representations, {ICLR} 2020, Addis Ababa, Ethiopia, April 26-30, 2020. OpenReview.net (2020), \url{https://openreview.net/forum?id=SkeHuCVFDr}

\end{thebibliography}

\appendix

\end{document}